\documentclass[11pt, a4paper, logo]{deepseek}

\usepackage[numbers, square, sort&compress]{natbib}
\usepackage{dblfloatfix}
\usepackage[export]{adjustbox}
\usepackage{ulem}
\usepackage{caption}
\usepackage{xspace}
\usepackage{pifont}
\usepackage{multirow}
\usepackage{makecell}
\usepackage{amsmath}
\usepackage{amssymb}
\usepackage{setspace}

\definecolor{citecolor}{HTML}{1976D2}
\hypersetup{
    colorlinks=true,
    linkcolor=black,
    filecolor=magenta,
    urlcolor=cyan,
    citecolor=citecolor,
}

\renewcommand{\today}{}

\title{\centering
    Remembering with Foresight: Pseudo-Query Guided Memory\\
    for Streaming Video Understanding
}
\vspace{-1cm}

\author{%
  \itshape\normalsize
  Shuning Wang\textsuperscript{1,2,$\ast$}, Zhiheng Wu\textsuperscript{1,$\ast$,$\dagger$}, Naiming Liu\textsuperscript{3}, YiNuo LU\textsuperscript{3},\par
  Bowen Liu\textsuperscript{4}, Xiao Tan\textsuperscript{1}, Kunbin Chen\textsuperscript{1}, Wei He\textsuperscript{1}, Yumeng Zhang\textsuperscript{1,$\ddagger$}\par
  \vspace{3pt}
  \normalfont\footnotesize\upshape
  \textsuperscript{1}Baidu Inc.\quad
  \textsuperscript{2}Zhejiang University\quad
  \textsuperscript{3}Harbin Institute of Technology\par
  \vspace{1pt}
  \textsuperscript{4}The Hong Kong University of Science and Technology\par
  \vspace{2pt}
  {\scriptsize
  $\ast$ Equal contribution.\quad
  $\dagger$ Project leader.\quad
  $\ddagger$ Corresponding author.}
}
\begin{abstract}
Recent advances in Multimodal Large Language Models have significantly improved video understanding. However, in streaming scenarios, existing methods struggle to retain the information critical for subsequent question answering under a limited memory budget, since future queries are unknown at the time of memory construction. Moreover, textual and visual memories are typically constructed independently, leading to redundancy and insufficient cross-modal coverage. In this paper, we propose \textbf{PQ-MeM}, a pseudo-query guided memory framework for streaming video understanding under strict capacity constraints. PQ-MeM explicitly models potential future information needs by constructing pseudo-queries, transforming passive memory into anticipatory memory. Specifically, these pseudo-queries are generated by integrating the training query distribution with the current scene context, thereby guiding memory construction. Based on this mechanism, we design a visual–textual bimodal memory structure that preserves high-level semantics in text and fine-grained visual evidence. Experimental results on multiple streaming and long-video benchmarks demonstrate that, under the same token budget, PQ-MeM consistently outperforms competitive baselines while maintaining low-latency responses required for real-time interaction.
\end{abstract}

\begin{document}

\maketitle
\vspace{-14pt}

\begin{center}
  \includegraphics[width=0.39\linewidth]{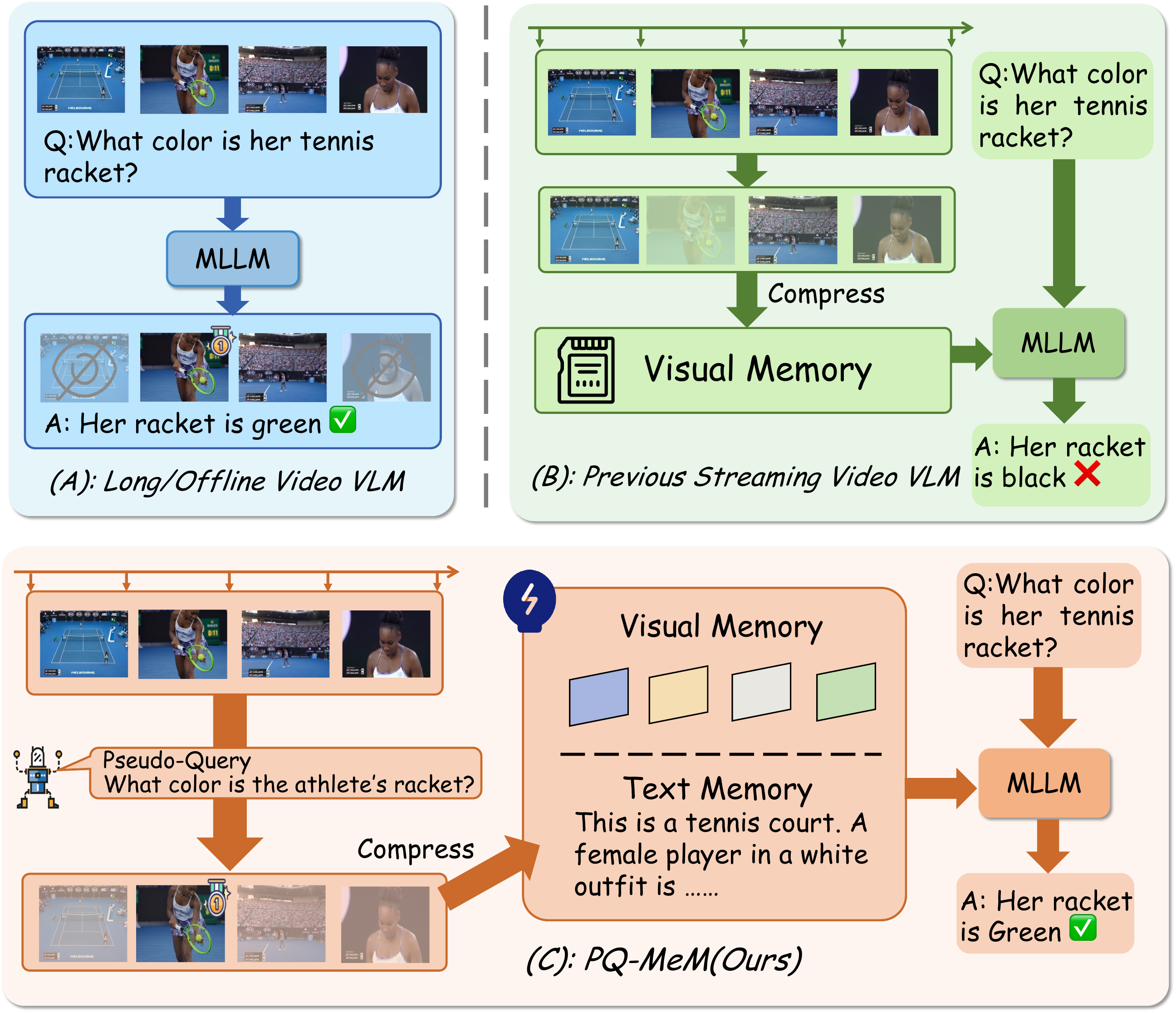}
\end{center}
\vspace{-12pt}
{\small
\captionof{figure}{\textbf{Comparison with previous methods.} (A) Offline Video VLM: key frames are selected based on query inputs. (B) Previous Streaming Video VLM: video features are passively compressed based on similarity. (C) PQ-MeM(Ours): a visual-textual dual-stream memory, actively selecting video features using pseudo-queries. The clarity of a visual memory reflects the visual feature representation: the clearer the frame, the more detail retained.}
\label{fig:intro}}

\newpage

\begin{spacing}{0.9}
\tableofcontents
\end{spacing}

\newpage

\section{Introduction}

In recent years, Multimodal Large Language Models (MLLMs) have achieved remarkable progress in video understanding tasks~\cite{bai2025qwen2, li2025videochat, song2024moviechat,yu2026minicpm}.
With the rise of real-world applications such as live streaming and embodied robots, research has gradually shifted from offline video understanding to the more challenging setting of streaming video processing~\cite{chen2024videollm,zhang2025flash,huang2025online,ning2025livevlm}. Unlike offline videos that are fully accessible before inference, streaming videos arrive incrementally, while queries often emerge after the relevant visual evidence has already passed. This mismatch creates an inherent dilemma: future queries are unknown, yet historical information cannot be retained indefinitely. Consequently, the core challenge lies in constructing compact memory representations that preserve long-term semantic information under strict capacity constraints.

Existing streaming video methods typically adopt passive memory retention strategies, such as uniform sampling~\cite{qian2024streaming, yao2025timechat} or fixed compression schemes~\cite{yang2025streammem,chen2026streamingtom}. The former often fails to preserve critical visual cues, while the latter may overlook important foreground content due to similarity-based aggregation. More importantly, these approaches lack explicit modeling of future information needs. Excessive retention rapidly exhausts the token budget, whereas aggressive compression may lead to the loss of evidence required for subsequent question answering. Essentially, they construct passive memory rather than anticipatory memory driven by future information demands. Therefore, the first key challenge is to model potential information needs when future queries remain unknown.

In memory construction, existing methods primarily rely on visual features for representation~\cite{zhao2026cogstream,wang2025videollamb}. However, visual tokens accumulate rapidly, requiring frequent compression and making it difficult to preserve key semantics over long video sequences. To address this limitation, recent works summarize streaming videos into compact textual memories to improve long-term contextual modeling~\cite{long2025seeing,chatterjee2025streaming}. Nevertheless, frames contain critical visual details potentially omitted from textual summaries. Moreover, textual and visual memories are often constructed independently~\cite{kang2025open}, with limited coordination, leading to redundancy and incomplete cross-modal coverage. Therefore, the second key challenge is coordinating textual
and visual memories to reduce redundancy and preserve
complementary high-level semantics and fine-grained visual evidence.


To address these challenges, we propose PQ-MeM (Fig.~\ref{fig:intro}), a pseudo-query guided memory framework for streaming video understanding under strict capacity constraints. Our key insight is to guide memory construction with potential future questions, enabling anticipatory retention before actual queries arrive. Specifically, we combine query-space anchors derived from training questions with scene-conditioned pseudo-queries to cover diverse potential information needs. Inspired by query-based keyframe selection in offline long-video understanding~\cite{zhang2025q,tang2025adaptive,zhao2025video}, we use these pseudo-queries to guide visual frame selection and compression, preserving evidence for subsequent question answering. We further coordinate textual and visual memories through pseudo-query guidance to reduce redundancy and preserve complementary high-level semantics and fine-grained visual evidence. Extensive experiments demonstrate that PQ-MeM surpasses its online baselines by 8.0 (Qwen2-VL) and 7.5 (Qwen2.5-VL) points on StreamingBench (Real-Time), with consistent gains on LVBench, supporting the effectiveness of the proposed memory framework for streaming video understanding.

Our main contributions are summarized as follows:


 
\begin{itemize}
\item We introduce the modeling of future information demands into streaming video understanding.
By constructing pseudo-queries to approximate the latent query space, we shift streaming memory construction from compression-driven passive retention to a demand-driven anticipatory paradigm.

\item  We propose PQ-MeM, a streaming memory framework that builds a complementary and query-aligned visual–textual memory under strict capacity constraints, enabling the collaborative preservation of high-level semantics and fine-grained visual evidence.
\item  Extensive experiments on streaming video benchmarks demonstrate consistent improvements in both real-time and long-horizon settings, validating the effectiveness and generalization of anticipatory memory construction.
\end{itemize}

\section{Related Work}
\textbf{MLLMs for Streaming Video Understanding.}
MLLMs achieve strong performance on image-level understanding and reasoning by aligning visual representations with large language models~\cite{yu2026minicpm,chen2024internvl,bai2025qwen2}. Recent studies extend them to the video domain~\cite{li2025videochat,tang2025video,chen2026streamingtom,xie2026fluxmem}, 
with representative works such as Video-ChatGPT~\cite{maaz2024video}, Video-LLaVA~\cite{lin2024video}, and 
TimeChat~\cite{ren2024timechat} adapting multimodal alignment and instruction tuning to general-purpose video understanding.
However, real-time streaming video poses a fundamental challenge: continuously arriving visual streams quickly exceed storage, token, and computational budgets. Meanwhile, earlier task-specific approaches (e.g., action recognition or video segmentation) focus on low-level visual modeling rather than maintaining semantically meaningful long-range context~\cite{wang2020towards,zhang2016real,wang2024uni}.
To address this, recent works explore language-centric paradigms for streaming video understanding and memory construction~\cite{zhang2025flash,yang2025streammem,qian2025dispider}: VideoLLM-Online~\cite{chen2024videollm} supports streaming narration and action anticipation, VideoChat-Online~\cite{huang2025online} maintains a 
hierarchical multi-resolution visual memory, and M3-Agent~\cite{long2025seeing} builds textual semantic memory for high-level reasoning over long streams.
Nevertheless, selecting and retaining task-relevant information from continuous streams under strict budgets remains challenging for real-time VQA.

\noindent
\textbf{Pseudo-Query Application.}
Pseudo-query generation aims to approximate potential information needs by constructing synthetic queries when real user queries or explicit guidance are unavailable, providing signals for retrieval and representation learning~\cite{gospodinov2023doc2query,wang2022gpl,dai2022promptagator}. Early work in text retrieval, such as Doc2query~\cite{nogueira2019document}, generates queries for documents to improve generalization. This idea has since been extended to multimodal settings. For visual grounding, Pseudo-Q~\cite{jiang2022pseudo} synthesizes language queries when real queries are unavailable, providing query-level guidance that improves cross-modal alignment between visual regions and textual descriptions. For text-to-video retrieval, Hybrid-Tower~\cite{lan2025hybrid} constructs pseudo-queries to strengthen video–text matching. However, in streaming video understanding, how to incorporate such query-level signals into memory construction so that stored information remains relevant to future queries is underexplored.


\section{Method}
\subsection{Overview}
\begin{figure}[t]
  \centering
  \includegraphics[width=0.92\linewidth]{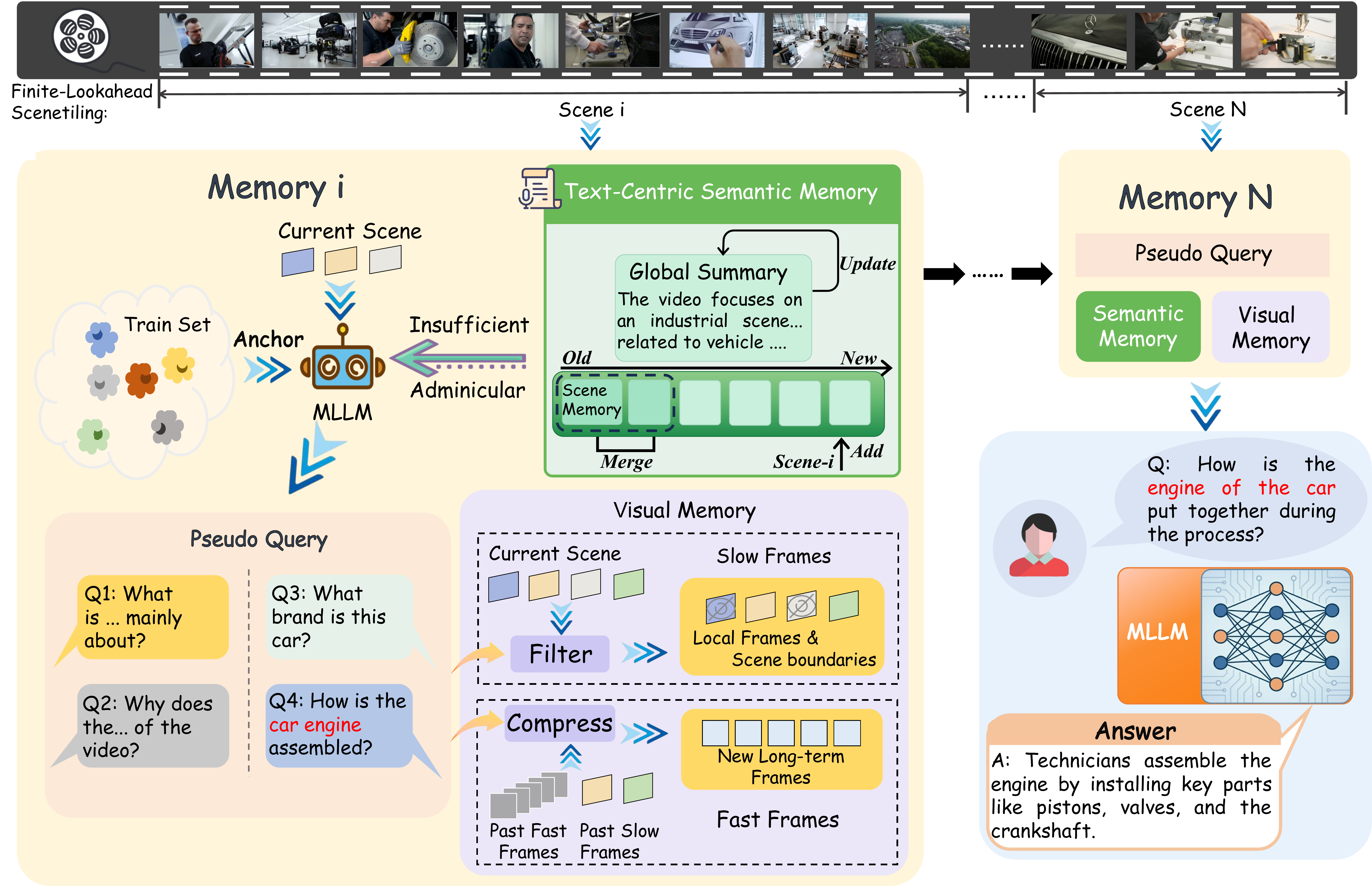}
  \caption{\textbf{Overview of PQ-MeM.} Finite-lookahead SceneTiling segments the stream and updates a text-centric semantic memory under a fixed budget. Query-space anchors and Scene frames generate pseudo-queries to guide visual memory, retaining key slow frames and compressing long-term fast frames. The pseudo-query Q4 anticipates the future query and preserves the evidence for answering.}
  \label{fig:framework}
\end{figure}

Streaming video understanding requires maintaining long-term
memory from observed frames without access to future user
queries~\cite{zhang2025flash,yang2025streammem}. 
To this end, we propose \textbf{PQ-MeM}, a \underline{P}seudo-\underline{Q}uery guided streaming \underline{Mem}ory framework that anticipates potential information needs through anchor-based pseudo-queries. As shown in Fig.~\ref{fig:framework}, PQ-MeM consists of three key components: (1) an online text-centric semantic memory that updates global and scene summaries in real time under a fixed token budget; (2) a future-oriented pseudo-query generation module that combines query-space anchors with scene-aware context to approximate the distribution of future queries; (3) a pseudo-query guided visual memory that selects slow (high-resolution) frames from the current scene and compresses fast (low-resolution) frames from past scenes. Together, these components enable compact, future query-aligned memory construction for streaming video understanding.

\subsection{Text-Centric Semantic Memory}

\noindent
To support long-term memory construction in streaming video understanding, we propose a Text-Centric Semantic Memory that converts visual observations into textual representations, which are organized through hierarchical consolidation under a fixed token budget. Scene-level memory is incrementally consolidated at the segment level, while global memory is continuously updated to maintain a compact semantic representation throughout streaming.

\noindent
\textbf{Finite-Lookahead SceneTiling.} SceneTiling identifies scene transitions by comparing similarity patterns in a video sequence, but it is primarily designed for offline video understanding~\cite{wang2025videollamb}. Its right-side similarity term $S_{ri}$ depends on future frames and is thus not directly applicable in the streaming setting. Therefore, we use a finite-lookahead sliding window to access short-term future frames and detect boundaries through changes in local consistency between consecutive frames.

Local consistency is measured using cosine similarity $S$, and scene boundaries are determined based on the depth score $d_i$, computed as follows:
\begin{equation}
d_i = \frac{S_{li} + \tilde{S}_{ri} - 2S_i}{2},
\label{eq:depth}
\end{equation}
where $S_i$ denotes the similarity between adjacent frames, $S_{li}$ is the maximum similarity within the past context, 
and $\tilde{S}_{ri}$ is the local right peak value calculated within a finite-lookahead window from $i$ to $i+\omega$:
\begin{equation}
\tilde{S}_{ri} = \max_{j \in (i,\; i+\omega]} S_j.
\label{eq:rightpeak}
\end{equation}
At query time, the model uses stored memory and buffered frames without
waiting for boundary confirmation or triggering compression. To detect scene boundaries online, we compare the depth score $d_i$ with an adaptive threshold $thr$ and declare a boundary at index $i$ when $d_i > thr$. The $thr$ is computed from the distribution of depth scores within a sliding past buffer:
\begin{equation}
thr = \mu_d + \alpha \sigma_d, 
\label{eq:threshold}
\end{equation}
where $\mu_d$ and $\sigma_d$ denote the mean and standard deviation of depth scores computed over the observed past frames, and $\alpha=0.7$.


\noindent
\textbf{Long-Horizon Streaming Textual Memory.} As shown in Fig.~\ref{fig:framework}, a long-horizon streaming textual memory is maintained to encode long-term semantics.
The memory consists of a Global Memory, which maintains a continuously updated summary, and a Scene Memory, which records recent scene segments in fine detail. During inference, relevant entries are retrieved for reasoning. To handle unbounded video length under a fixed token budget, we keep a bounded episodic window and periodically merge older entries using an MLLM, preserving key semantics while enabling constant-memory streaming.

\subsection{Future-Oriented Pseudo-Query Generation}

In streaming settings, the absence of future queries makes memory construction inherently uncertain. Saliency-based or uniform compression methods often fail to align stored content with future queries~\cite{zhang2016real}. 
We address this by introducing anchor-based pseudo-queries that simulate future questions to guide memory construction. To provide effective query-level guidance, the pseudo-queries are designed to cover a diverse range of information needs. Additionally, these pseudo-queries should be answerable only through visual information, encouraging the memory to retain visually relevant details that are useful for unseen queries.

\noindent
\textbf{Query-Space Anchors for Pseudo-Queries.} 
To obtain representative anchors that broadly cover the query embedding space, we embed all training questions into a text embedding space and select $K$ of them as anchors via k-center optimization~\cite{gonzalez1985clustering}.
Specifically, each question is encoded by the CLIP~\cite{radford2021learning} text encoder, forming a candidate set:
\begin{equation}
\mathcal{U} = \{ u_1, u_2, \ldots, u_M \}, 
\quad u_i \in \mathbb{R}^d,
\end{equation}
where $u_i$ denotes the embedding of the $i$-th question.
The $K$ anchor queries are obtained by
\begin{equation}
{A}^* =
\operatorname*{argmin}_{\mathcal{C} \subset \mathcal{U}, \, |\mathcal{C}| = K}
\max_{u \in \mathcal{U}}
\min_{c \in \mathcal{C}}
\|u - c\|_2.
\end{equation}
The resulting ${A}^*$ denotes the selected set of $K$ query-space anchors, $\mathcal{C}$ is a candidate subset of $\mathcal{U}$ with size $K$, and $c \in \mathcal{C}$ denotes an individual anchor. This objective minimizes the maximum distance from any question to its nearest anchor, yielding $K$ well-distributed anchors ${A}^*$ that cover the query embedding space.

\noindent
\textbf{Scene-Aware Pseudo-Query Generation.}
We then introduce a scene-level pseudo-query generator conditioned on the query-space anchors, producing scene-aware pseudo-queries that approximate future information needs.
For each scene $k$ with visual frames $v_k$ and textual summary $t_k$, the generator is guided by the anchor set ${A}^*$ to produce a set of pseudo-queries:
\begin{equation}
\tilde{Q}_k = \{q_{k,1}, \ldots, q_{k,N}\} = \mathcal{G}(v_k, t_k, {A}^*),
\end{equation}
where $\tilde{Q}_k$ denotes the pseudo-query set associated with the scene, and $\mathcal{G}$ represents the pseudo-query generator instantiated by Qwen2.5-VL-3B~\cite{bai2025qwen2}. To ensure the pseudo-query guided visual memory complements rather than duplicates the text-centric memory, $\mathcal{G}$ is instructed to generate pseudo-queries answerable only from visual cues in $v_k$, not from the textual summary $t_k$.
This is designed to steer the answers to $\tilde{Q}_k$ toward evidence beyond $t_k$, thereby biasing subsequent visual memory selection toward information complementary to the text-centric memory.


We unify the query-space anchors and the scene-aware pseudo-queries to form the final pseudo-query set $Q_k = {A}^* \cup \tilde{Q}_k$. The query-space anchors cover diverse information needs, while the scene-aware pseudo-queries align memory with visual evidence.
This strategy provides structured guidance for memory construction, directing retention and compression toward broadly useful and visually complementary information.

\subsection{Pseudo-Query Guided Visual Memory}
\begin{figure}[h]
  \centering
  \includegraphics[width=0.68\linewidth]{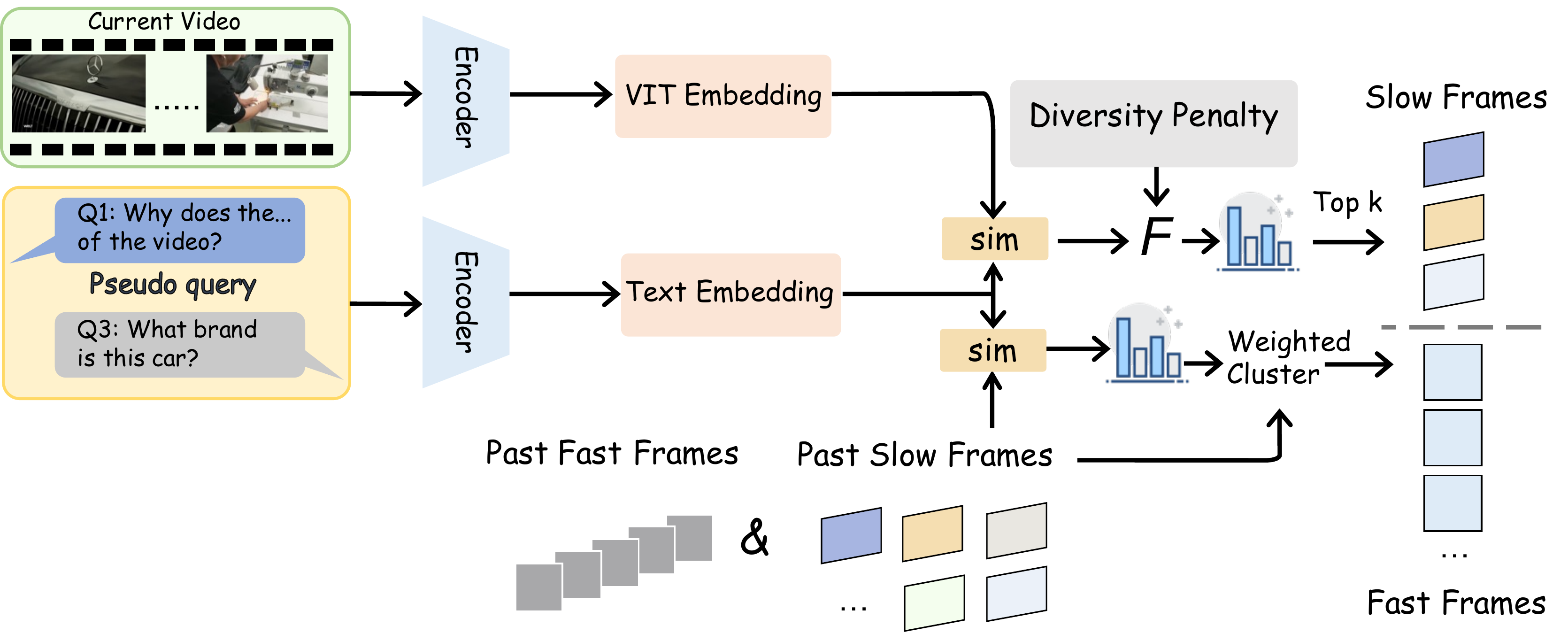}
  \caption{\textbf{Illustration of pseudo-query guided visual memory construction.} Pseudo-queries drive the filtering and compression of slow frames and fast frames, ensuring that important visual evidence is retained while reducing redundancy for efficient streaming video understanding.}
  \label{fig:visual_memory}
\end{figure}

We propose pseudo-query guided visual memory construction for streaming video understanding. Under a dual-memory design~\cite{yang2025kwai}, slow-frame memory preserves high-resolution evidence while fast-frame memory stores compressed representations. Unlike query-agnostic methods, our approach uses pseudo-queries to guide frame retention and compression, as illustrated in Fig.~\ref{fig:visual_memory}, prioritizing information that is useful for future queries and complementary to the text memory. Each frame $V_t$ at time step $t$ is encoded into a visual representation $e_t$ by the visual encoder $\mathcal{E}_{\text{enc}}(\cdot)$:
\begin{equation}
e_t = \mathcal{E}_{\text{enc}}(V_t).
\end{equation}

\noindent
\textbf{Pseudo-Query Guided Slow-Frame Filter.} In streaming video understanding, certain future queries rely on a few frames that capture critical events, and the frames around scene boundaries carry substantial semantic information. Therefore, both types of frames are preserved as high-resolution slow frames, so that this evidence is not discarded during compression.
To limit memory cost, we select $N_{\text{slow}}$ slow frames from the current scene and the scene-boundary candidates. 
This selection is driven by a scoring function $\Phi_{\text{slow}}(f)$ that combines a marginal pseudo-query coverage gain with a Maximal Marginal Relevance (MMR) term~\cite{agarwal2010maximum}, so that the highest-scoring frames are retained while redundant ones are penalized:

\begin{equation}
\Phi_{\text{slow}}(f) = \lambda \cdot \Delta(f \mid \mathcal{X}) + (1 - \lambda) \cdot \mathrm{MMR}(f \mid \mathcal{X}),
\end{equation}
where $\mathcal{X}$ is the set of frames selected so far, initialized as $\mathcal{X}=\varnothing$ and updated after each selection, and $\lambda$ trades the coverage gain against the $\mathrm{MMR}$ term. The marginal coverage gain $\Delta(f \mid \mathcal{X})$ of a candidate frame $f$ is defined as
\begin{equation}
\Delta(f \mid \mathcal{X})
=
\frac{1}{\lvert Q_k \rvert}
\sum_{m=1}^{\lvert Q_k \rvert}
\max \big( 0,\; S_{f,m} - C_m(\mathcal{X}) \big).
\end{equation}
Here, $S_{f,m}$ denotes the cosine similarity between frame $f$ and query $m$, and $C_m(\mathcal{X})$ is the maximum similarity between query $m$ and any frame in $\mathcal{X}$. MMR is calculated as:
\begin{equation}
\mathrm{MMR}(f \mid \mathcal{X})
=
\beta \cdot \mathcal{R}(f)
-
(1-\beta)\cdot \mathcal{D}(f \mid \mathcal{X}),
\end{equation}
where $\mathcal{R}(f)$ denotes the average similarity between frame $f$ and all pseudo-queries, and $\beta$ balances relevance $\mathcal{R}(f)$ against redundancy $\mathcal{D}(f \mid \mathcal{X})$. Here, $\mathcal{D}(f \mid \mathcal{X})$ measures the redundancy of frame $f$ with respect to the selected frame set $\mathcal{X}$, defined as:
\begin{equation}
\mathcal{D}(f \mid \mathcal{X})
=
\max_{f' \in \mathcal{X}}
\frac{e_f^{\top} e_{f'}}
{\|e_f\| \, \|e_{f'}\|},
\end{equation}
where $e_f$ and $e_{f'}$ are the feature vectors of frames $f$ and $f'$, and $\| \cdot \|$ denotes the Euclidean norm of the vectors. Frames are greedily selected to maximize the marginal coverage gain with MMR regularization until $N_{\text{slow}}$ frames are obtained.

\noindent
\textbf{Pseudo-Query Guided Fast-Frame Compression.}
To efficiently summarize long-term frames, a pseudo-query guided importance weighting scheme is introduced to compress visual information, combining global query relevance and difficulty-aware contribution.
Global relevance $H_f$ is:
\begin{equation}
H_f = \sum_{m=1}^{|Q_k|} S_{f,m},
\end{equation}
where $S_{f,m}$ measures the similarity between frame $f$ and query $m$, reflecting how broadly the frame supports the pseudo-query set.
Let $\mathcal{F}$ denote the compressed fast-frame set, and $f'$ denote the fast-frame, The coverage $\Psi_m(\mathcal{F})$ of query $m$ is
\begin{equation}
\Psi_m(\mathcal{F})
=
\frac{1}{|\mathcal{F}|}
\sum_{f' \in \mathcal{F}} S_{f',m},
\end{equation}
which captures how well the query has already been supported. Query difficulty $D_m(\mathcal{F})$ is defined as
\begin{equation}
D_m(\mathcal{F})
=
1 -
\frac{\Psi_m(\mathcal{F})}
{\max S_{f,m}},
\end{equation}
so that insufficiently covered queries receive larger weights. 
The difficulty-aware contribution $G_f$ of frame $f$ is
\begin{equation}
G_f = \sum_{m=1}^{|Q_k|} S_{f,m}D_m(\mathcal{F}),
\end{equation}
which emphasizes frames that help improve poorly covered queries.
The overall importance score $\Phi_f$ is
\begin{equation}
\Phi_f
=
\theta \cdot H_f + (1-\theta)\cdot G_f,
\end{equation}
where $\theta$ balances global relevance and difficulty-aware gain. The weights $\Phi_f$ are injected into the k-means~\cite{mcqueen1967some} objective as sample importance weights~\cite{zhang2025flash}, allowing the clustering process to prioritize query-relevant frames.

\section{Experiments}
\subsection{Experimental Settings}
\textbf{Datasets.} We fine-tune all backbones on a 13k-sample subset of LLaVA-Video-178K~\cite{zhang2024llava}, a large-scale video instruction dataset covering tasks such as captioning, open-ended VQA, and multiple-choice VQA. The questions in this subset also form the training query pool from which the query-space anchors are drawn, so that the anticipated information needs are grounded in a real video QA distribution.

\noindent
\textbf{Benchmark.} We conduct multiple-choice VQA experiments on six benchmarks, none of which contributes to our fine-tuning data: StreamingBench~\cite{lin2024streamingbench}, OVO-Bench~\cite{niu2025ovo}, and OV-Bench~\cite{huang2025online} for the streaming setting, where questions arrive while the video is still unfolding; LVBench~\cite{wang2025lvbench} (average duration over 1 hour) and Video-MME~\cite{fu2025video} (11 seconds to one hour) for long-video memory retention; and MVBench~\cite{li2024mvbench} as the short-video reference. For the offline benchmarks, each question is delivered only after the stream reaches the end of the video, so answers may rely exclusively on the memory accumulated causally so far.


\begin{table}[t]
\centering
\setlength{\tabcolsep}{6pt}
\renewcommand{\arraystretch}{1.15}
\caption{Performance comparison with SOTA video MLLMs on long and streaming-video benchmarks. All our results are obtained under streaming constraints with a 12.7k-token budget. $\Delta$ reports the improvement over each corresponding online baseline under identical token usage.}
\begin{adjustbox}{max width=1.0\textwidth}
\begin{tabular}{l c c c c c c c c}
\toprule
\multirow{3}{*}{Method} & 
\multirow{3}{*}{Base Model} & 
\multirow{3}{*}{Size} & 
\multicolumn{1}{c}{StreamingBench} & 
\multicolumn{1}{c}{OVOBench} & 
\multicolumn{1}{c}{OV-Bench} & 
\multicolumn{1}{c}{VideoMME} & 
\multicolumn{1}{c}{LVBench} & 
\multicolumn{1}{c}{MVBench} \\
\cmidrule(lr){4-6} 
\cmidrule(lr){7-8} 
\cmidrule(lr){9-9}
& & & Real-Time & All & Overall & w/o sub & Avg & Avg \\
\midrule
\rowcolor{blue!7}
\multicolumn{9}{c}{\textit{Open-source Offline Video MLLMs}} \\
\hline
InternVL2~\cite{chen2024internvl}        & -- & 8B & 63.7 & 50.1 & 48.7 & 54.0 & -- & 65.8  \\
Qwen2-VL~\cite{wang2024qwen2}            & --     & 7B & 69.0 & 52.7 & 49.7 & 58.3 & 41.4 & 65.1  \\
LongVA~\cite{zhang2024long}              & --    & 7B & 60.0 & --   & 43.6 & 52.6 & -- & --  \\
LLaVA-OneVision~\cite{li2024llava}       & --    & 7B & 70.1 & 52.9 & 49.5 & 58.2 & -- & 58.6  \\
Video-LLaMA2~\cite{zhang2023video}       & --    & 7B & 49.2 & --   & --   & 47.9 & -- & 54.6  \\
LLaVA-Video~\cite{zhang2024llava}        & --   & 7B & --   & 53.1 & 53.0 & 63.6 & -- & 58.6 \\
\midrule
\rowcolor{blue!7}
\multicolumn{9}{c}{\textit{Open-source Online Video MLLMs}} \\
\hline
VideoLLM-online~\cite{chen2024videollm}  & LLaMA2      & 8B & 36.0 & 12.8 & 9.6  & 26.9 & 24.0 & 33.9  \\
MovieChat~\cite{song2024moviechat}       & Vicuna     & 7B & --   & --   & 30.9 & 38.2 & --   & 55.1  \\
VideoChat-Online~\cite{huang2025online}  & InternLM2     & 4B & --   & --   & 54.9 & 52.8 & --   & 64.9  \\
Dispider~\cite{qian2025dispider}         & Qwen2    & 7B & 67.6 & 41.8 & --   & 57.2 & --   & --  \\
Flash-VStream~\cite{zhang2025flash}      & Qwen2     & 7B & 67.0 & 52.6 & 47.4 & 58.3 & 39.2 & 64.3  \\
StreamingTOM~\cite{chen2026streamingtom}     & LLaVA-OV      & 7B & -- & -- & -- & 59.9 & -- & --  \\
FluxMem~\cite{xie2026fluxmem}      & Qwen2.5     & 7B & 76.4 & 53.3 & -- & 65.3 & -- & --  \\
Qwen2-VL-Online & -- &  7B & 62.3 & 52.6 & 48.4 & 57.3 & 38.8 & 63.0 \\
\rowcolor{green!15}
PQ-MeM(Qwen2-VL)                         & Qwen2  & 7B & 70.3 & 58.3 & 51.2 & 60.7 & 42.3 & 65.8  \\
$\Delta$(vs Qwen2-VL-Online)   & -- & -- & \textbf{+8.0} & \textbf{+5.7} & \textbf{+2.8} & \textbf{+3.4} & \textbf{+3.5} & \textbf{+2.8} \\
Qwen2.5-VL-Online & -- & 7B & 70.4 & 57.3 & 51.5 & 59.4 & 40.6 & 65.7 \\ 
\rowcolor{green!15}
PQ-MeM(Qwen2.5-VL)   & Qwen2.5  & 7B & \textbf{77.9} & \textbf{60.8} & \textbf{57.1} & 63.8 & \textbf{44.6} & \textbf{66.9}  \\
$\Delta$(vs Qwen2.5-VL-Online) & -- & -- & \textbf{+7.5} & \textbf{+3.5} & \textbf{+5.6} & \textbf{+4.4} & \textbf{+4.0} & \textbf{+1.2} \\

\bottomrule
\end{tabular}
\end{adjustbox}
\label{tab:main_results}
\end{table}

\noindent
\textbf{Implementation Details.} 
We adopt Qwen2-VL~\cite{wang2024qwen2} and Qwen2.5-VL~\cite{bai2025qwen2} as backbones, both initialized from their public checkpoints and retaining the original ViT-based visual front end with dynamic-resolution tokenization and spatial token merging.
During evaluation, videos are sampled at 1 fps, and the memory is capped at a budget of roughly 12.7k tokens per video. For a fair comparison, each backbone is also evaluated by sampling frames uniformly over the whole stream under the same 12.7k-token budget, referred to as Qwen2-VL-Online and Qwen2.5-VL-Online, respectively. Online baselines use public checkpoints; the same LoRA tuning applied to them changes their results by at most $\pm0.6$ points.
Training is performed with LoRA instruction tuning, freezing the visual encoder and adapting the projector and LLM linear layers. 
All experiments are conducted on 8 NVIDIA H20 GPUs. The visual memory hyperparameters $\lambda, \beta, \theta$ are set to $\{0.9, 0.7, 0.6\}$ with 5 anchors and 5 scene-generated queries per scene.

\subsection{Main Results}
In this section, we compare our PQ-MeM method with the following state-of-the-art (SOTA) methods. These include SOTA models for long-video understanding such as Qwen2.5-VL~\cite{bai2025qwen2}, LongVA~\cite{zhang2024long}, LLaVA-OneVision~\cite{li2024llava}, and LLaVA-Video~\cite{zhang2024llava}, as well as for streaming video understanding, including MovieChat~\cite{song2024moviechat}, VideoChat-Online~\cite{huang2025online}, Dispider~\cite{qian2025dispider}, Flash-VStream~\cite{zhang2025flash}, StreamingTOM~\cite{chen2026streamingtom}, and FluxMem~\cite{xie2026fluxmem}.

PQ-MeM improves all six benchmarks across both backbones, with gains reaching $+8.0$ on StreamingBench, $+5.7$ on OVOBench, and $+4.0$ on LVBench over the corresponding online baselines (Table~\ref{tab:main_results}). The strongest configuration (Qwen2.5-VL) reaches 77.9, 60.8, and 44.6 on StreamingBench, OVOBench, and LVBench, exceeding every other entry in Table~\ref{tab:main_results}. Notably, even the Qwen2-VL variant surpasses its offline counterpart (70.3 vs.\ 69.0 on StreamingBench and 42.3 vs.\ 41.4 on LVBench) despite operating under a 12.7k-token memory budget. The smallest gain, $+1.2$, falls on the nearly saturated MVBench. Since the frame rate, model size, and token budget are identical to those of the online baselines, these improvements are attributable to the memory design rather than to additional capacity.

\begin{table}[t]
\setlength{\tabcolsep}{8pt}
\centering
\caption{\textbf{Ablation study on the Fast-Slow visual memory.} This experiment is conducted with Qwen2-VL as the backbone and evaluates the contribution of the scene slow-frame memory (SF), boundary slow-frame memory (BF), and long-term fast-frame memory (LFF). The row marked ``uni.'' replaces the LFF with uniform sampling.}
{
\begin{tabular}{ccc cccc}
\toprule
SF & BF & LFF & \makecell{Streaming\\Bench} & OVOBench & VideoMME & Avg \\
\midrule
\texttimes & \checkmark & \checkmark & 59.9 & 51.0 & 54.2 & 55.0\\
\checkmark & \texttimes & \checkmark & 63.8 & 53.8 & 56.2 & 57.9\\
\checkmark & \checkmark & uni. & 65.4 & 56.3 & 56.7 & 59.5\\
\checkmark & \checkmark & \checkmark & 70.3 & 58.3 & 60.7 & 63.1\\
\bottomrule
\end{tabular}
}
\label{tab:visual_memory1}
\end{table}


\begin{table}[t]
\centering
\setlength{\tabcolsep}{6pt}
\caption{\textbf{Ablation study on pseudo-query guidance.}
All variants use Qwen2-VL and the same 12.7k-token memory budget.}
\begin{tabular}{lccc}
\toprule
Method & StreamingBench & OVOBench & VideoMME \\
\midrule
w/o guidance           & 66.1 & 54.9 & 58.3 \\
w/ scene-fact guidance & 67.7 & 55.2 & 59.3 \\
w/ pseudo-query (ours) & \textbf{70.3} & \textbf{58.3} & \textbf{60.7} \\
\bottomrule
\end{tabular}
\label{tab:train_component}
\end{table}


\subsection{Ablation Study}
\noindent
\textbf{Ablation Study on Fast-Slow Visual Memory.} Our visual memory module consists of three components: a \emph{scene slow-frame memory} (SF) that selects key frames from the current scene via pseudo-query, a \emph{boundary slow-frame memory} (BF) sampled around scene boundaries to preserve key transitions, and a \emph{long-term fast-frame memory} (LFF) that compresses past frames via pseudo-query clustering.
To validate the Fast-Slow design, we ablate each component by removing it or replacing LFF with a uniform baseline (uni.).
As shown in Table~\ref{tab:visual_memory1}, removing SF or BF causes drops on all three benchmarks, while replacing LFF with uniform compression also degrades performance (Avg: 59.5 vs. 63.1). The full model achieves the best results, indicating that both SF and BF contribute to short-term evidence retention, and that LFF is crucial for long-horizon information under streaming constraints.


\noindent\textbf{Ablation Study on Pseudo-Query Guidance.}
We compare two ways to guide visual memory construction: describing visual facts versus generating questions likely to be asked later (Table~\ref{tab:train_component}). Relative to unguided memory, scene facts improve accuracy by 1.6, 0.3, and 1.0 points on StreamingBench, OVOBench, and VideoMME. Pseudo-queries further outperform scene facts by 2.6 points on StreamingBench, 3.1 on OVOBench, and 1.4 on VideoMME. Thus, across all three benchmarks, guiding memory with potential questions yields higher QA accuracy than guiding it with descriptions of the same scene.

\noindent\textbf{Ablation Study on the Text--Visual Bimodal Memory.}
We ablate the two memory streams and the visual--textual exclusivity constraint under the same 12.7k-token budget, where each variant removes one stream and reallocates its tokens to the other, so that the comparison reflects how the budget is partitioned rather than how much of it is used (Table~\ref{tab:bimodal}). Removing the textual memory and granting its tokens to the visual stream lowers LVBench from 42.3\% to 40.7\%, while removing the visual memory instead lowers it to 36.5\%, so neither stream reproduces the full model even when granted the entire budget. Removing the visual--textual exclusivity constraint of Sec. 3.3 reduces accuracy from 42.3\% to 41.1\%, suggesting that targeting details omitted from the summary benefits QA.
\begin{table}[!b]
\centering
\setlength{\tabcolsep}{6pt}
\caption{Text--visual bimodal memory ablation on LVBench (Qwen2-VL backbone).
All variants share the same total token budget. ``Exclusivity'' refers to the visual--textual exclusivity constraint of Sec.~3.3, which requires pseudo-queries to be answerable only from visual cues rather than from the textual summary.}
\begin{tabular}{lc}
\toprule
Variant &  LVBench \\
\midrule
w/o text memory               & 40.7 \\
w/o visual memory             & 36.5 \\
w/o visual--textual exclusivity & 41.1 \\
Full PQ-MeM                   & \textbf{42.3} \\
\bottomrule
\end{tabular}

\label{tab:bimodal}
\end{table}



\subsection{Pseudo-Query Alignment and Generalization}
We evaluate how well pseudo-queries align with future questions and how training-derived anchors and scene-generated queries contribute to QA accuracy. On LVBench, $69.1\%$ of ground-truth questions match at least one pseudo-query at a cosine similarity threshold of $0.7$. Without pseudo-query guidance, the model achieves $39.4\%$ accuracy. Using anchors alone raises accuracy to $40.6\%$, while scene-generated queries alone reach $41.2\%$. Combining both yields $42.3\%$. Thus, scene-generated queries improve accuracy both with and without anchors, by $1.7$ and $1.8$ percentage points, respectively, while anchors contribute a further $1.1$ points on top of scene-generated queries. We also evaluate PQ-MeM on ActivityNet-QA~\cite{yu2019activitynet}, an open-ended QA benchmark held out from training. With Qwen2-VL-7B as the backbone, PQ-MeM achieves 61.7\% accuracy, improving by 4.3 points over our reproduced baseline of 57.4\% under the same streaming protocol and token budget.

\begin{figure}[t]
  \centering
  \includegraphics[width=0.52\linewidth]{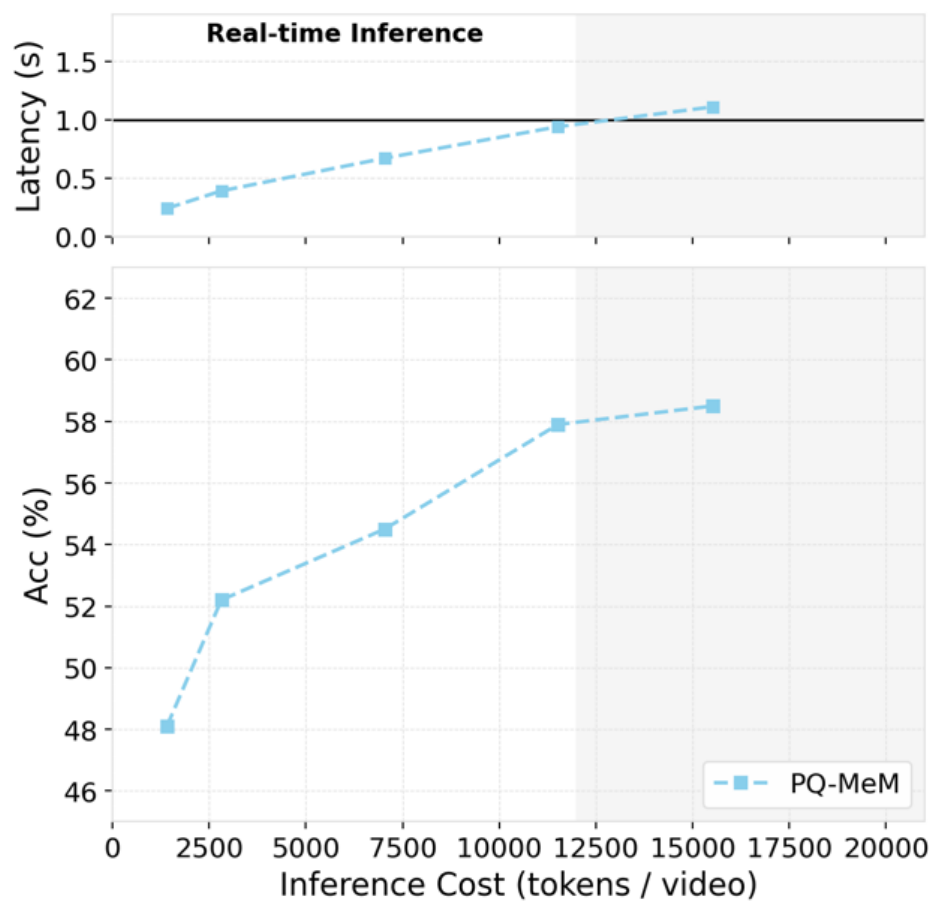}
  \caption{\textbf{Computational Efficiency.} Query-time response latency (top) and accuracy (bottom) on VideoMME-medium under varying token budgets.}
  \label{fig:ablation_window_a}
\end{figure}

\subsection{Computational Efficiency}
Figure~\ref{fig:ablation_window_a} reports query-time latency and QA accuracy on the medium-length subset of VideoMME as we vary the maximum number of video tokens used for inference. We use a single H20 GPU with BFloat16 precision and FlashAttention-2. At the 12.7k-token budget used in our main experiments, PQ-MeM handles a question in under 1\,s. Computing pseudo-query similarities takes approximately 100\,ms, but this occurs during idle streaming periods before a query arrives and is therefore excluded from the query-time latency. Pseudo-query generation runs asynchronously at scene boundaries and completes between successive boundaries in our experiments, so its runtime is excluded from the reported query-time latency.


\subsection{Hyperparameter Sensitivity Analysis}
\begin{figure}[!h]
  \centering
  \includegraphics[width=1.0\linewidth]{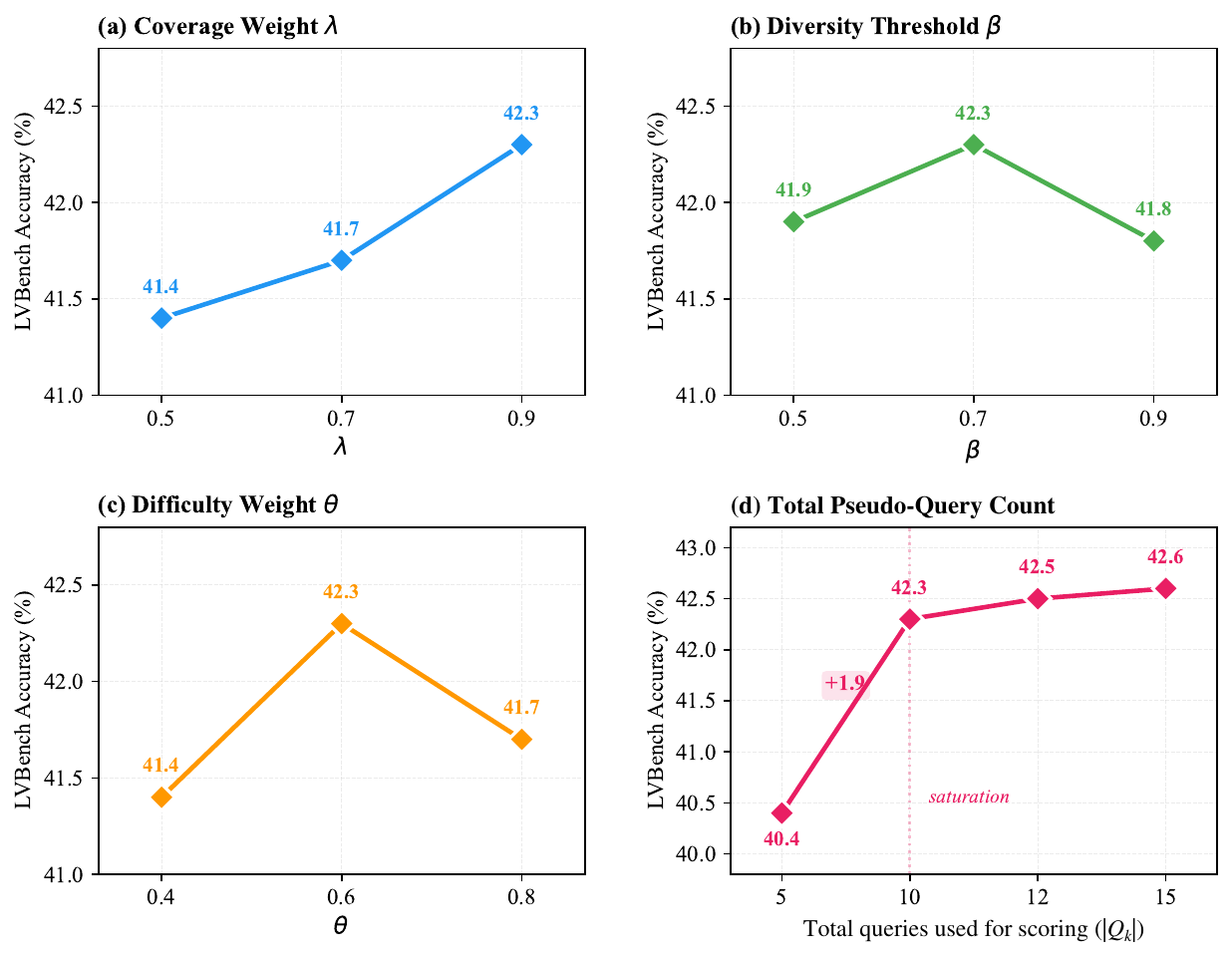}
\caption{\textbf{Hyperparameter sensitivity analysis on LVBench with Qwen2-VL as backbone.} Accuracy under varying $\lambda$, $\beta$, $\theta$, and total query count $\lvert Q_k\rvert$, with all other hyperparameters fixed.}
  \label{fig:sens}
\end{figure}
As shown in Fig.~\ref{fig:sens}, varying any one of the three weights ($\lambda$, $\beta$, or $\theta$) changes accuracy by at most 0.9 percentage points over the tested range. Accuracy increases with $\lambda$ and peaks at $\beta=0.7$ and $\theta=0.6$ in their respective sweeps. Increasing the pseudo-query count from 5 to 10 improves accuracy from 40.4\% to 42.3\%, while increasing it from 10 to 15 yields only 0.3 additional percentage
points. We therefore use 10 pseudo-queries and fix $(\lambda,\beta,\theta)=(0.9,0.7,0.6)$ across all benchmarks in Table~\ref{tab:main_results}.

\subsection{Case Study}
\begin{figure}[!t]
  \centering
  \includegraphics[width=0.72\linewidth]{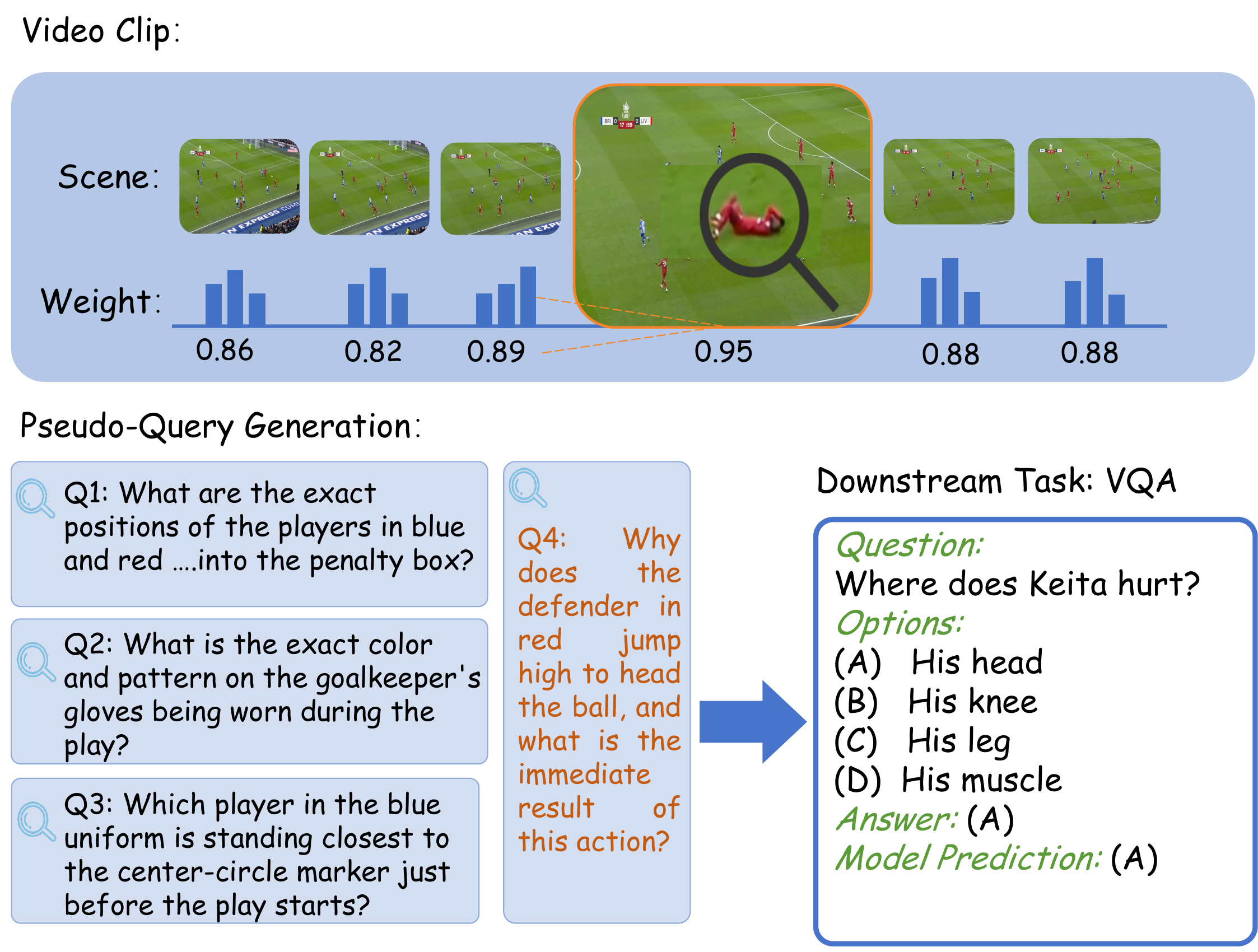}
  \caption{\textbf{Case study.} Pseudo-query Q4 demonstrates strong semantic alignment with the future VQA question, leading to a substantially higher importance weight assigned to the corresponding key frame. }
  \label{fig:case_study}
\end{figure}

We present a case study of pseudo-queries in long-term memory compression (Fig.~\ref{fig:case_study}). Pseudo-query Q4 concerns the event relevant to the future VQA question. Under pseudo-query guidance, the corresponding key frame receives the highest importance weight (0.95), compared with 0.82–0.89 for the other illustrated frames. This example illustrates how pseudo-query guidance helps preserve potentially useful visual evidence under a constrained token budget.

\newpage
\section{Conclusion}
In this paper, we presented PQ-MeM, a pseudo-query guided memory framework that transforms passive memory into anticipatory memory for streaming video understanding. By approximating future query needs with pseudo-queries, PQ-MeM enables proactive, query-aligned memory construction under limited capacity. We further introduced a visual-textual bimodal memory to preserve both long-term semantics and fine-grained visual evidence. Experiments demonstrate consistent improvements over strong baselines across streaming benchmarks while maintaining low-latency responses. Future work will extend PQ-MeM to broader streaming vision tasks and open-ended interactive settings.

{
    \small
    \bibliography{main}
}

\end{document}